# Empowering remittance management in the digitised landscape: A real-time Data-Driven Decision Support with predictive abilities for financial transactions

**Full research paper**

**Rashikala Weerawarna**
Newcastle Business School
The University of Newcastle
New South Wales, Australia
Email: Rashikala.Weerawarna@uon.edu.au

**Shah J Miah**
Newcastle Business School
The University of Newcastle
New South Wales, Australia
Email: Shah.Miah@newcastle.edu.au

## Abstract

The advent of Blockchain technology (BT) revolutionised the way remittance transactions are recorded. Banks and remittance organisations have shown a growing interest in exploring blockchain's potential advantages over traditional practices. This paper presents a data-driven predictive decision support approach as an innovative artefact designed for the blockchain-oriented remittance industry. Employing a theory-generating Design Science Research (DSR) approach, we have uncovered the emergence of predictive capabilities driven by transactional big data. The artefact integrates predictive analytics and Machine Learning (ML) to enable real-time remittance monitoring, empowering management decision-makers to address challenges in the uncertain digitised landscape of blockchain-oriented remittance companies. Bridging the gap between theory and practice, this research not only enhances the security of the remittance ecosystem but also lays the foundation for future predictive decision support solutions, extending the potential of predictive analytics to other domains. Additionally, the generated theory from the artifact's implementation enriches the DSR approach and fosters grounded and stakeholder theory development in the information systems domain.

**Keywords**: Remittance, Blockchain, Machine Learning, Big Data analytics, Design Science Research

## 1 Introduction

The global remittance industry plays a vital role in the movement of funds across borders, facilitating transactions for millions of individuals and businesses. Issues in complex cross-border remittance processors include regulatory compliance challenges, high cost, slow processing time, a lack of transparency, and low security. These issues hinder the efficiency, affordability, and trustworthiness of the remittance processors that are used in the traditional environment. BT emerges as the preferred choice for remittance over other technologies due to its distinctive attributes. Its decentralised architecture ensures the security and integrity of transaction data required by remittance processors. Unauthorised alternations, reduced security, less transparency, high costs, and efficiency issues in traditional remittance can be eliminated by BT. Moreover, by eliminating intermediaries and streamlining cross-border remittances, blockchain minimises the costs and accelerates transaction speed. This unique combination of security and transparency makes blockchain a compelling choice for improving the remittance process, setting it apart from other technology solutions. While blockchain is gaining popularity in the finance field as a secure, transparent, tamper-proof and cost-effective intermediatory-free platform, cross-border remittance management faces numerous challenges and opportunities. As the volume of transactions and complexity increases, the need for efficient monitoring systems becomes paramount. This research aims to address the need by proposing a data-driven predictive DSS for real-time monitoring of blockchain-oriented remittance transactions. Moreover, it serves as a valuable artefact for remittance managers, empowering them with the capability to efficiently identify risks and detect transaction anomalies. By incorporating ML algorithms, the artefact will proficiently identify anomalies and potential risks, thereby strengthening fraud prevention measures and enhancing decision making processes. The foundation of the study lies in the theory generating DSR approach (Perrers et al. 2007; Beck et al. 2023) which aims to address real-world problems in remittance monitoring by creating an innovative artefact and validating it through rigorous methods. Furthermore, this enables scholars to possess the valuable ability to extract theoretical insights from analytical abstractions. Concurrently, the proposed data-driven DSS represents an artefact that aligns with the theoretical contributions to the stakeholder's theory, design science theory and grounded theory. The theoretical contribution of the study is threefold. Firstly, it exemplifies how the predictive decision support approach, a type of data-driven DSS (as identified by Power 2002 and 2007) embraces stakeholder centric approach leading to a more robust solution that maximises value for stakeholders in the remittance industry. Secondly, it enriches DSR by demonstrating how the practical application of theory-generating research can effectively address real-world challenges in the remittance industry. Furthermore, this study contributes to grounded theory by validating the efficiency of the proposed data-driven DSS through real-world case studies, providing its efficiency in mitigating risks and enhancing transaction monitoring in the digitised landscape.

This paper presents a comprehensive exploration of the research framework, begins a literature review that investigates remittance transactions, BT, data-driven DSS, and the integration of ML for big data predictive analytics. Subsequently, the research methodology is described, including, the data-driven DSS artefact, data collection, analytical process and evaluation. It discusses the primary findings and concludes by addressing limitations and future research directions.

## 2 Literature Review

BT is reshaping various conventional sectors, including the finance services industry. Traditional banking systems typically operate remittance processors using centralised architecture, while the engagement of multiple intermediaries in global remittance operations elevates expenses of cross-border payments. Furthermore, the immense potential of BT has captured significant attention from banks and financial organisations. The existing research reveals that banks are continuously exploring possibilities to integrate blockchain solutions. Several initiatives like Ripple leverage, BT to establish alternative cross-border remittance infrastructure (Gupta 2018, Weerawarna et al. 2023) This innovation has the capacity to fundamentally reshape the framework of the traditional remittance system for various reasons. Blockchain introduces a decentralised and transparent ledger system, where transactions are recorded and validated across multiple nodes without the need for a central authority's consensus. The main reasons are the ability to reduce costs by eliminating intermediaries and streamlining the remittance process. Traditional systems often involve multiple banks and payment processors, resulting in higher fees and slower transactions. Blockchain's peer-to-peer model cuts out these middlemen, making remittances more affordable and efficient. Its decentralised, tamper-proof nature ensures data integrity and security. Immutability and transparent records foster trust and prevent unauthorised changes, enhance transaction reliability, have faster processing and reduce the risk of errors through automation and smart contracts, blockchain provides a seamless customer experience (Guo and Liang 2016). However, implementing blockchain in remittance systems faces challenges, including cryptocurrency volatility, use education, interoperability, lack of standardisation and technological complexities (Zhang et al. 2020; Weerawarna et al. 2023). Despite these obstacles, blockchain's prominent attributes establish it as a secure, efficient, and cost-effective alternative to traditional remittance systems, ensuring a transparent and trustworthy remittance platform.

The literature on remittance transactions embraces a broad range of studies exploring their significance in the global economy. In the context of remittance, complying with AML regulations necessitates obtaining specific information, including metadata, through the Know Your Customer (KYC) process (Battistini, 2016). According to AUSTRAC (Australian Transaction Reports and Analysis Centre), both the KYC process and the identification of beneficial owners are deemed essential elements in remittance. BT has gained significant attention from scholars in recent years due to its potential to disrupt various industries including business, finance and remittance (Chong et al. 2019). The decentralised nature of BT and its role in enhancing security, transparency, and trust in financial transactions highlighted the benefits of real-time, low-cost and transactional immutability to accelerate the implementation of BT in banking and remittance industries. The integration of BT holds significant promise for transforming traditional remittance practices in the digital era and managing them involves critical business processes such as KYC and Anti-Money Laundering (AML) monitoring. In particular, transaction control and monitoring play a fundamental role in ensuring compliance in the remittance industry. The emergence of the digitising era provides the remittance industry with a data-driven DSS to tackle transaction monitoring insights. This involves the constant monitoring of transactional level data in real time.

Big data has been considered the next frontier for real-time innovations and its usage is expected to become a crucial factor for transactional level forecasting and decision making in the digitised future of remittance. The financial big data generated by blockchain-oriented remittance transactions enables stakeholders to interpret insights for monitoring and assessing the AML processors (Yoo 2017). Real-time transaction monitoring, especially when analysed using ML techniques generates substantial business value in the remittance industry. ML algorithms applied to monitor remittance transactions can effectively detect patterns indicative of fraudulent activities. By learning from historical data, ML models can identify suspicious transaction patterns, unusual behaviour, or anomalies that may signify fraud attempts. This proactive approach empowers remittance businesses to prevent fraud, safeguard customer funds, and uphold service integrity. Furthermore, ML analysis of transaction data facilitates the identification of high-risk transactions, the assessment of money laundering or terrorist financing probabilities, and the flagging of potential risks. Consequently, businesses can implement risk mitigation measures, such as enhanced due diligence or transaction verification, to minimize financial and reputational risks. Moreover, ML-driven transaction monitoring enhances compliance by promptly identifying suspicious transactions and non-compliant behaviour, thereby enabling timely reporting and risk mitigation.

Numerous studies have explored the implementation of blockchain solutions in various domains. For instance, Miyachi et al. (2021) introduced a hybrid framework that combines on-chain and off-chain mechanisms to preserve privacy in healthcare data management. Søgaard et al. (2021) employed a DSR approach to develop a prototype platform for value-added tax settlement. Additionally, Wouda et al. (2019) developed a blockchain application aimed at enhancing the transaction process for office buildings in the Netherlands. In the financial sector, Soonduck (2017) highlighted the transformative potential of BT, particularly in the field of remittance. Furthermore, Lee (2022) conducted a study to examine whether FinTech and BT could offer solutions to mitigate risks in the banking industry, specifically concerning de-risking. While various blockchain solution design studies have been proposed across different areas within the finance sector (Yoo 2017), there remains a specific research gap concerning remittance management. This gap has been acknowledged by several researchers, including Soonduck (2017), and Emily (2022). The existence of this gap becomes evident when considering the existing studies on blockchain in finance, as shown in Table 1.

Previous research has predominantly focused on challenges in KYC process, transaction delays, lack of privacy, illegal activities in public blockchain, mining speed, privacy, user permission, transaction visibility issues in private blockchains ( Bhumika, et al., 2017). Researchers have suggested storing KYC data off the chain to address challenges in the traditional remittance process, which involves lengthy back-office and regulatory checks (Yadav et al. 2019; Malhotra et al. 2021). Integrating BT with big data and analytics, employing ML, can help overcome limitations in KYC optimization (Yadav et al 2019; Malhotra et al 2021).

The design of a blockchain-oriented transaction management application holds significant potential in leveraging both off-chained data (capturing users' behaviour) and on-chained data (smart contract data). The transaction data recorded on the blockchain offers structured, secure, and valuable information for big data analytics (Fedak, 2018). Adopting BT for remittance streamlined money transfer management that results in greater efficiency and transparency (Keller, 2018). Researchers have investigated the integration of data analytics, ML, and predictive modelling techniques into data-driven DSS to enhance decision-making capabilities (Li et al., 2022). Studies have also focused on credit risk assessment (Liu et al., 2022), fraud detection (Chang et al., 2022), and market risk analysis using machine learning techniques (Broby, 2022). The literature emphasizes the importance of big data in training robust predictive models and highlights the benefits of machine learning in enhancing risk assessment accuracy and efficiency.

## 2.1 Theoretical perspectives

This smart data-driven DSS development study involved the DSR approach, grounded and stakeholder theories in the cross-border remittance domain. The involvement of industry 4.0 technologies could offer tailored insights that empower decision making for managers in different fields. Data-driven DSSs have gained substantial recognition within the information systems research field due to their role in enhancing informed decision-making processes across many domains (Miah and McKay 2016; Borrero and Mariscal 2022; Jiskani et al. 2022; Yazdani et al. 2023; Unhelkar et al 2022). DSS are computerised information systems designed to aid operations and management by presenting business data in a user-friendly manner, enabling smoother business decision-making. As Power (2002) notes, DSS are tailor-made to streamline decision processes, prioritizing support for decision-makers rather than complete automation. Their agility to swiftly adapt to evolving decision-maker requirements is a defining characteristic. Integrating data analytics and stakeholder involvement in data-driven DSS bridge the technology and managerial decision making and addresses complex challenges across diverse domains (Gupta et al. 2022).

The nature of DSR, an iterative problem-solving approach allows researchers to identify challenges and requirements. DSR's emphasis on user involvement and feedback fosters collaboration with stakeholders, resulting in an intuitive data-driven DSS tailored to their needs. Stakeholders theory shaped the landscape of data-driven DSS research by integrating stakeholder perspectives into the fabric of effective DSS formation. Engagement of stakeholders in the remittance field not only enriches the practical utility of data-driven DSS but also fosters a sense of ownership among them leading to higher rates of adoption and effectiveness in the remittance field. Intelligent data-driven DSS has the potential to significantly enhance data accessibility in remittance and empower managers with valuable insights into organizational processes, customer behaviour, and comprehensive organization-wide performance metrics. The involvement of stakeholder theory in the development of an intelligent data-driven DSS represents a strategic and holistic approach that elevates DSSs from a technical solution to a collaborative endeavour. Moreover, stakeholder theory ensures the ethical considerations, privacy concerns of sensitive data and social responsibility that make data-driven DSS more effective technology innovation. The iterative development and refinement of the data-driven DSS foster long term stakeholder relationships which leads more robust digital landscape for the remittance field.

Grounded theory's core principle of deriving theory from empirical data aligns (Beck et al. 2023) perfectly with the dynamic and complex landscape of cross-border remittance. By immersing in real world data (Akoka et al. 2023) from remittance transactions, customer behaviour grounded theory guides the identification of patterns, relationships and insights that are directly applicable to the functionalities of the data-driven DSS. Further, the grounded theory emphasises constant comparison and validation of emerging trends in remittance transactions. This leads predictive analytics algorithms and decision support mechanisms to be rigorously benchmarked against real data. Iteratively, testing and refining these components make data-driven DSS robust and assist in addressing the precise needs of remittance managers and other stakeholders.

While existing literature provides valuable insights into the digital landscape, certain gaps remain to be addressed. Specifically, there is a lack of research that comprehensively integrates BT with data-driven DSS for real-time monitoring of remittance transactions while leveraging ML and predictive analytics for risk assessment. The literature lacks studies of challenges and opportunities associated with adopting ML in the context of blockchain-oriented remittance transactions. The current research aims to address these gaps by proposing a data-driven DSS that integrates ML and analytics to enhance real-time monitoring and financial risk assessment for remittance transactions.

*Table 1: Existing studies on blockchain in finance*

| The main point of the studies | Used analytical methods | Research methodologies | Problem gist | Authors |
|---|---|---|---|---|
| Automate KYC on blockchain | Natural Language Processing (NLP) | Systematic Review | Intelligent document scanning improves the client onboarding process | Malhotra et al. (2021) |
| KYC Optimization using blockchain Smart Contract Technology | Smart Contract | Design Research Methodology | Secured and cost-efficient KYC process. | Parra et al. (2017) |
| Autoregressive Distributed Lag (ARDL) model uses to examine the impacts of remittances on terrorism. | Autoregressive Distributed Lag (ARDL) | Used annual time series data, The unit root test, the co-integration technique, Hypothesis testing | Using annual time series data covering the period of 1990-2019 generated from the World Economy Index (2017), www.globaleconomy.com | Yilmaz et al. (2020) |
| Big data in accounting and BT in financial security and cybersecurity | Big data analytics | Literature review | By reviewing the literature that includes topics; Big datain Accounting, blockchain's use in financial security and cybersecurity and the Department of Homeland Security plan for cybersecurity. | Demirkan et al. (2020) |
| Blockchain affects the business intelligence efficiency of banks | The partial least square technique | A survey method | A survey method was used to collect data from banks of the Nanjing City. | Ji and Tia (2022) |

## 3 Research Methodology and Process

DSR has gained significant attention in the field of information systems research (Beinke et al. 2019; Miah et al. 2016, 2018, 2019; Ostern et al. 2021; Jardim et al. 2021) providing a systematic process for creating, improving, and evaluating IT artifacts. The initial DSR methodology aligns with this research goal, following the six-activity framework (Peffers et al., 2007). In this study, we adopt the theory generating DSR approach introduced by Beck et al. (2023), which combines DSR and grounded theory techniques to offer a valuable capability to derive theoretical insights from analytical abstractions. By incorporating elements from behavioural science and grounded theory methods, we construct theoretical foundations based on real data, addressing methodological gaps present in traditional DSR. The application of theory generating DSR assist in this study to advance our understanding of the design and valuation of IT artifacts and their impact on the real-world  Furthermore, it leads us to gain deeper insights and a higher level of analytical abstraction, enriching the theoretical contributions of DSR in information system research.

The theory-generating DSR approach is a suited methodology for this study as it aims to address real-world problems in the remittance industry through the creation of an artefact.  The approach empowers the development of data-driven DSS artefact by integrating theoretical framework with practical implementation. It focuses on problem solving, iterative refinement, theory integration and theoretical contribution that advance the knowledge in the field of decision support in cross-border remittance.

The development of a data-driven DSS for real-time monitoring of blockchain-oriented remittance transactions big data represents the artifact. The research approach involves iterative cycles of design, implementation, and evaluation as explained below, where each iteration enhances the artifact based on theoretical insights and empirical evidence (Figure 1).

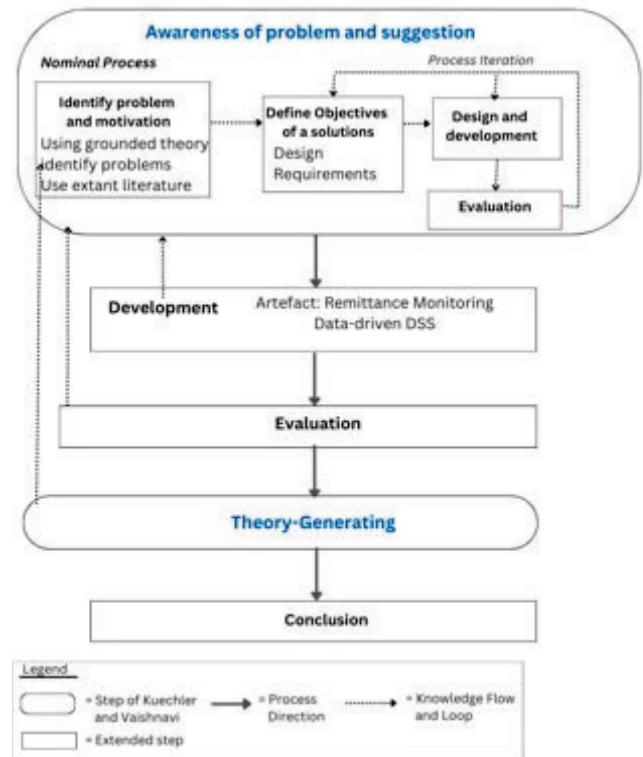

*Figure 1: Research process, adopted from theory generating DSR process (Beck et al. 2023)*

## 3.1 Awareness of problem and suggestion

With the rapid growth of cross-border fund transfers, there is a pressing need for efficient and secure monitoring systems to mitigate potential risks and fraudulent activities. There is a necessity to develop a digitised landscape for remittance managers that supports their decision making, refines risk identification protocols and elevates operational efficiency, regulators provide accurate valuable insights referring to transaction transparency and security. Therefore, suggesting an integration of BT and predictive analytics data-driven DSS. The proposed data-driven DSS represents a novel and innovative approach, addressing the current limitations in traditional remittance monitoring practices.

### 3.1.1 Theoretical sampling

We identify and analyse the real-world problem in remittance monitoring and determine the essential requirements and elements for the remittance monitoring artifact. We not only focused on the existing issues but also anticipated potential emergent problems in the remittance domain.

### 3.1.2 Slices of data

The data gathered and analysed relate to the process of remittance management and monitoring Initially, we conducted a literature analysis to explore the application of BT in cross-border remittance services from the management perspectives, then researched data analytics and ML on remittance monitoring. To gather the necessary requirements, we consulted with stakeholders in remittance organisation and a researcher specialising in blockchain regulatory requirements. Additionally, the insights gained from the literature review played a significant role in shaping the objectives and data requirements of the artifact.

In the cross-border remittance industry, various types of business data sets are utilised. These include transaction data, customer data, market and exchange rate data, and compliance and regulatory data. Transaction data comprises information about individual remittance transaction details such as the sender's and recipient's names, addresses, identification numbers, transaction amounts, currencies involved, transaction timestamps, and associated fees. Customer data includes information about the individuals or businesses involved in sending or receiving remittances. This data comprises personal details such as names, addresses, contact information and identification documents such as passports or ID cards. Additionally, customer data may include transaction history to track the remittance activities of customers. Market and exchange rate data provides information on foreign exchange rates, market trends, and economic factors that affect remittance. Compliance and regulatory data refer to the information required to comply with AML and KYC procedures that involve ensuring regulatory compliance, conducting customer due diligence, assessing risks, and fulfilling reporting obligations as per relevant laws and regulations. This research utilises remittance transaction data to gain valuable insights into cross-border remittance businesses as mentioned in Table 2.

*Table 2: Research data*

| Transaction Data |
|---|
| sender name; sender addresses; sender identification numbers; receiver name, receiver; addresses; receiver;identification numbers; transaction amounts; transaction reason; currency involved; transaction timestamps; associated fees |
| transaction frequency; transaction history |

In the context of this research, it is important to address ethical considerations, particularly pertaining to data privacy and security measures. Given the sensitive nature of cross-border transactions, the protection of individuals' data and the assurance of a secure monitoring process are necessary. By acknowledging and adhering to robust data privacy and security measures, this study not only aims to enhance remittance monitoring efficiency but also sustain ethical standards, ensuring the protection and privacy of sensitive information.

## 4  Proposed artefact and design requirements

While Figure 1 illustrates the process of this study Figure 2 depicts the dynamic interaction between key technological elements in the study: blockchain, big data, ML, and data analytics, all working in synergy to develop a DSS to have better remittance monitoring. The process commences with the initiation of remittance transactions by the blockchain. During the execution of a remittance, the system leverages blockchain-driven transaction data to train the ML model for real-time pattern recognition and predictions. This model is utilised to classify transactions based on the attributes of both the sender and the receiver.

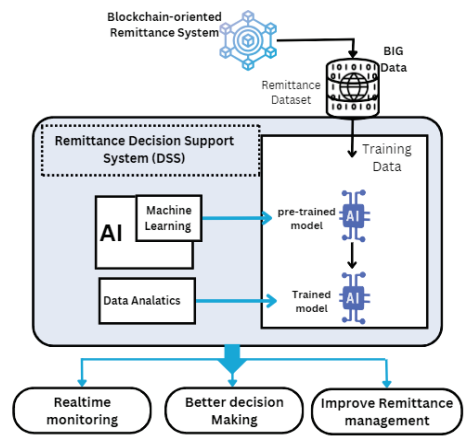

*Figure 2: Design construct of proposed DSS*

Necessary transaction data related to cross-border remittances can be extracted using blockchain explorers (such as Etherscan.io for Ethereum or Blockchair.com which supports multiple blockchains) or Application Programming Interfaces (APIs) specific to the Remittance Blockchain such as Ethereum API. In this research, we collect remittance data using an Ethereum test bed simulation by setting up an Ethereum network. We defined the parameters for simulated transactions such as sender and recipient addresses, amounts, and currencies. Execute these transactions within the simulation to generate data, recording details like transaction hashes, sender and recipient addresses, transaction timestamps, transaction values, gas fees, transaction types, block information such as block height, block timestamp, block size, and the number of transactions in a block, network metrics such as hash rate, difficulty level, block propagation time, or transaction throughput, user attributes such as wallet addresses, transaction history, or user-provided metadata, token-related data that includes token balances, token transfer events, token supply, token distribution, or token transaction volumes, and time series data like transaction volumes over time, transaction value fluctuations, or transaction frequency. While the simulation may not fully replicate the actual Ethereum blockchain, this approach provides a controlled environment to generate valuable data for analysis and experimentation.

In the ML model selection and training step concentrate on choosing appropriate ML algorithms based on the specific analysis goals and available data. By employing the ML algorithms and PD techniques suggested below, the data-driven DSS can provide remittance managers with real-time insights, enabling them to make informed decisions, mitigate financial risks, and enhance transaction monitoring in the blockchain-oriented remittance ecosystem.

Logistic regression and gradient boosting are valuable ML algorithms used in this research.

Logistic regression is where the goal is often to classify transactions as either legitimate or potentially fraudulent, excel in binary classification tasks and accurately distinguish legitimate from suspicious transactions. It serves as a robust tool for predictive analysis and risk assessment and integration into real-time monitoring enhances prompt action. Its logistic Regression's adaptability, evaluation metrics, and threshold adjustment enable it to capture evolving fraud patterns effectively.

Gradient boosting produces a robust predictive model that excels in identifying complicated fraud patterns. It offers high accuracy, crucial for spotting anomalies in cross-border transactions. Furthermore, it handles imbalanced data, a common scenario in remittance monitoring, and can effectively prioritise minority class samples during training. Additionally, it provides insights into feature importance, aiding the identification of key factors influencing fraud. The technique's adaptability, scalability, and real-time capabilities make it ideal for rapid decisions and timely fraud detection. By aggregating predictions from diverse models, gradient boosting mitigates overfitting risks. Its continuous learning and model calibration abilities equip data-driven DSS to address evolving fraud tactics and tailor its sensitivity to specific requirements.

In the context of this research, which focuses on blockchain data and employs logistic regression for fraud detection, the primary evaluation metrics of interest include accuracy, precision, recall, F1 score, Area Under the ROC Curve (AUC-ROC) and Area Under the Precision-Recall Curve (AUC-PR). When applying regression-based models to analyse blockchain data, such as predicting transaction volumes, the relevant metrics are Mean Squared Error (MSE) and Root Mean Squared Error (RMSE). Accuracy was suitable to predict classes correctly. These chosen metrics play a crucial role in striking the right balance between minimising false positives and maximising the detection of actual fraudulent transactions highlighting the precision-recall trade-off is a vital consideration when dealing with imbalanced blockchain datasets. Below, we explain the ML models and their corresponding metrics.

- Logistic Regression: To model binary outcomes and assess the likelihood of customers being high-risk or low-risk based on historical transactional data and associated risk indicators. As metrics precision and recall assist, assess its suitability to predict high-risk customers. AUC-ROC and AUC provide an overall view of its performance in fraud detection.
- Random Forest: To build an ensemble of decision trees, offering improved accuracy and robustness in identifying potential risk customers through feature importance analysis. Precision, recall and F1 score are used to evaluate, its ability to identify risk customers.
- Gradient Boosting: To create a strong predictive model by iteratively combining weak learners, improving the accuracy of risk assessment and anomaly detection. AUC-ROC and AUC-PR measure overall performance and imbalanced data handling.
- Support Vector Machines: To classify customers based on their risk levels using a hyperplane in a high-dimensional feature space, aiding in distinguishing risk categories. Precision and recall metrics evaluate and classify customers based on risk categories.
- Predictive analytics techniques such as anomaly detection and clustering will be integrated into the data-driven DSS to identify suspicious transactions and segment customers based on their risk profiles. In this precision, recall and F1 score metrics evaluate anomalies.
- Time series forecasting models such as Autoregressive Integrated Moving Average. Mean absolute error, mean squared error, root mean squared error, mean absolute percentage error metrics provide insight into forecasting accuracy.

The following text outlines how we plan to interpret the results. In logistic regression, we choose accuracy when our research objective is to predict classes correctly, and the classes are reasonably balanced. Accuracy is a straightforward metric that measures the overall correctness of predictions. A high accuracy score indicates that a significant portion of the data is classified correctly. Precision and recall are selected when the dataset is imbalanced, and we need to manage false positives (precision) or false negatives (recall). These metrics help us understand the trade-off between minimising different types of errors. Precision focuses on the percentage of true positive predictions among all positive predictions, while recall measures the percentage of true positives found among all actual positives. The interpretation of these metrics depends on the relative importance of false positives and false negatives in the context of the problem. The F1-Score is chosen when balancing false positives and false negatives is crucial. AUC-ROC is appropriate when we want to assess the overall classification performance and understand the trade-off between true positives and false positives at different classification thresholds. The AUC-ROC quantifies the model's ability to discriminate between positive and negative instances. A higher AUC value indicates better performance. It can also help determine the optimal classification threshold for the problem. In gradient boosting, we choose MSE for regression tasks when the research objective is to minimise the squared errors between predicted and actual values. RMSE is also used as Lower RMSE values indicate better model performance, and like MSE, it is sensitive to outliers. A lower MSE indicates that the model's predictions are closer to the actual values. A higher AUC-ROC score will be used to indicate better discrimination between positive and negative instances, with 0.5 indicating random performance and 1.0 indicating perfect performance. Data-driven DSS offers an array of features that empower remittance monitoring managers with enhanced decision-making capabilities. These features, as outlined by Power (2007), cater to users' needs for systematic data exploration, insightful visualisations, and efficient data management.

- Ad hoc data filtering and retrieval: Remittance managers can systematically search and retrieve computerised data through user-friendly interfaces. Dropdown menus and predefined queries facilitate filtering, while drill-down capabilities enable detailed analysis.
- Alerts and triggers: This data drives users to set rules for notifications, enabling timely alerts for specific events such as potential fraud. Managers can establish email notifications or predefined actions to ensure quick responses to critical situations.
- Data displays: Remittance managers can choose from various displays, such as scatter diagrams, and pie charts. Interactive options allow users to modify displays aiding trend analysis.
- Data management: Allows users to work with a subset of data within working storage.
- Data summarisation: The custom aggregations and calculated field summarises allow flexible perspectives on the data.
- Metadata creation and retrieval: Remittance managers can enhance analyses and reports by adding metadata.
- Report design and generation: Managers can design and present formal reports with infographics.
- Statistical analysis: Descriptive statistics, trend lines, and data mining capabilities enable users to extract insights from the data, supporting informed decision-making.
- Predefined data displays: Operational performance monitoring remittance transactions use dashboard displays. This feature offers easy access for users, enhancing their ability to review standardised insights.

By leveraging these features, a well-designed data-driven DSS equips remittance monitoring managers to access accurate, reliable, and high-quality information, leading to informed and timely decisions. The DSS ensures a single version of the truth, supports individual analyses, and strengthens the overall decision-making process. Depending on the nature of the analysis, the above suggested options may be used for data analysis and then split the pre-processed data into training and testing sets. Finally, train the ML model on the training data, allowing it to learn the patterns and relationships between the features and the target variable such as transaction volume, transaction success, failure and trends. The suggested data-driven DSS comprises the following components:

- Data Collection Layer which gathers real-time remittance transaction data from blockchain - oriented remittance networks.

- Data Processing Layer: In this layer, the raw data is pre-processed, cleaned, and transformed into structured datasets suitable for analysis.
- ML Layer: The heart of the data-driven DSS lies in the ML layer, where predictive analytics models are integrated. This layer helps to analyse the structured data to identify patterns, detect anomalies, and classify customers based on their risk levels.
- Visualization and Reporting Layer: An interactive and intuitive interface for remittance managers to visualize the analysed data, risk assessments, and transactional insights. Real-time dashboards and reports are generated, presenting the identified risk customers and suspicious transactional activities.

## 5   Theorical influence and evaluation

The evaluation step trains the ML model using the testing dataset to assess its performance and generalisation ability. In this step we measure the model's performance using appropriate evaluation metrics based on the analysis goal, such as mean absolute error (MAE), root mean squared error (RMSE), accuracy, or precision and recall. Here, we can validate the model's predictive capabilities by comparing its predictions against the actual cross-border remittance transaction data.

Theory generation facilitates the discovery of additional theoretical insights that go beyond the scope of the developed artefact, highlighting an interconnected process of problem-solving and theorising. Additional theoretical sampling offers valuable new insights into how the artefact is used and performed. To achieve this, additional data collection is conducted after data-driven DSS creation, with the support of grounded theory. The purpose of this is to explore various aspects of the data-driven DSS, including its performance, usability, and incorporation. Additional data are sourced from relevant knowledge in existing literature and findings derived from the artefact evaluation. This step in the research elevation to a higher conceptual level as it extends beyond mere literature and encompasses diverse elements such as prototypes and innovative data. Our research process starts with open coding, grouping indicators from data or initial IT artefacts into concepts and categories over time, and focusing on themes of central interest for theoretical insights.

The emerged category in the previous step allows for analyses of the relationships among them and thus increases insights. The result adds a contribution to the remittance industry in the form of grounded theory about the developed artefact. This expands the knowledge base that contributes to stakeholders' theory design science theory and grounded theory. After identifying and defining the core categories, we assess the relationships among them to generate additional theoretical insights. An important step to achieve the final theoretical contribution of this study entails extensive comparisons across the generated insights and prior published work in the same domain. The emergent contribution of our research and its theoretical insights then can be integrated into follow-up DSR projects as new requirements to consider.

## 6. Discussion and Conclusion

Through rigorous evaluation and validation, the research aims to describe a new data-driven DSS artefact and demonstrate the effectiveness and practicality of the solution. It plays a vital role in remittance monitoring by analysing data comprehensively, enabling real-time oversight, predictive insights, and pattern recognition. Customisable dashboards offer instant views, while automated alerts ensure timely responses. Its data-backed insights empower informed decisions, adapting to evolving fraud tactics. The DSS aids compliance and strategic planning, enhancing risk management and decision-making for remittance managers. The research makes several notable contributions to the field of data-driven DSS and blockchain-oriented remittance transactions. The data-driven DSS artifact will empower remittance managers with real-time insights into transactional activities, enabling proactive risk mitigation and improved decision-making.

Ultimately, this research endeavours to bridge the gap between theoretical advancements and practical implementations in the financial domain, providing valuable insights for academia, industry, and policymakers. Further, it contributes to financial organisations and their understanding of how BT can be utilized for remittance purposes. The project aids stakeholders in the finance sector in improving their knowledge of blockchain-oriented remittance, and its opportunities, and ultimately using blockchain remittance analytics as a tool to understand financial transaction behaviours and predict the preferences of transaction holders. Considering the involvement of multiple stakeholders in the artifact's usability, regulatory requirements, technical requirements, and the social impact and mindset related to blockchain remittance, further improvements can be made with valuable contributions to stakeholders' theory. The same model can be tested and developed using different smart contract deployment platforms such as Hyperledger Fabric, Corda, Stella, and others. Furthermore, this model can serve as a case study opportunity for blockchain research in other disciplines such as healthcare, education, supply chain management, IoT, and more (de Vass, Shee, and Miah, 2018). Future research in this domain contains the potential to enhance our understanding of the complex relationships between data-driven DSS (Ali and Miah, 2017) and other aspects such as BT and the multifaceted landscape of remittance transactions. One avenue for research involves the refinement of evaluation metrics (e.g. through a design study (Miah et al. 2019)). User experience and usability evaluations, coupled with feedback, can be enhanced further. Regulatory implications and behavioural analytics provide avenues to explore broader effects and real-world impact of this advancement. Each of these directions contributes to an enriched and comprehensive understanding of blockchain-oriented remittance transactions and data-driven DSS, enhancing the decision-making capabilities of financial stakeholders and policymakers while unlocking the technology potential in various industries.

## Copyright